\pdfoutput=1
\documentclass{article}

\usepackage{arxiv}

\usepackage[utf8]{inputenc}
\usepackage[T1]{fontenc}
\usepackage{amsmath}
\usepackage{amssymb}
\usepackage{hyperref}
\usepackage{url}
\usepackage{booktabs}
\usepackage{amsfonts}
\usepackage{nicefrac}
\usepackage{microtype}
\usepackage{graphicx}
\usepackage[numbers,sort&compress]{natbib}
\usepackage{doi}

\usepackage{xcolor}
\usepackage{multirow}
\usepackage{enumitem}
\usepackage{tikz}
\usetikzlibrary{positioning, arrows.meta, fit, backgrounds, calc, shapes.geometric}
\usepackage{xspace}

\newcommand{\bench}{\textsc{RAGStress}\xspace}

\title{\textsc{RAGStress}: A Controlled Benchmark for Evaluating Retrieval-Augmented Generation Under Knowledge-Base Degradation}

\author{
  Shiqi Yang\thanks{This work does not relate to the author's position at Amazon.} \\
  Amazon \\
  \texttt{sallyyan@amazon.com} \\
  \And
  Jiekai Ma\footnotemark[1] \\
  Amazon \\
  \texttt{jiekal@amazon.com} \\
  \And
  Gaoyuan Du\footnotemark[1] \\
  Amazon \\
  \texttt{gdu@amazon.com} \\
}

\renewcommand{\shorttitle}{\textsc{RAGStress}: Evaluating RAG Under Knowledge-Base Degradation}

\hypersetup{
  pdftitle={RAGStress: A Controlled Benchmark for Evaluating Retrieval-Augmented Generation Under Knowledge-Base Degradation},
  pdfauthor={Shiqi Yang, Jiekai Ma, Gaoyuan Du},
}

\begin{document}
\maketitle

\begin{abstract}
Retrieval-Augmented Generation (RAG) is typically evaluated under the implicit
assumption that the underlying knowledge base (KB) is clean, leaving the
behaviour of RAG systems under realistic KB degradation poorly characterised.
We introduce \bench, a controlled evaluation benchmark for
stress-testing RAG systems under systematic KB corruption. The benchmark pairs
four naturalistic corruption types (factual corruption, numeric typo,
relevance poisoning, contradiction injection) with three severity levels
(subtle, moderate, obvious) over a single-KB, metadata-filtered experimental
design built from 57 MMLU subjects and 182{,}546 documents.
Across 52{,}500 model--question--condition evaluations, \bench
reveals that clean retrieval can mask robustness differences,
semantic-fidelity corruptions are substantially more harmful than
signal-utility perturbations, no-retrieval accuracy does not predict
corrupted-retrieval robustness, and mixed-KB accuracy should not be treated
as worst-case robustness. We document the benchmark's intended use, supported
claims, and limitations, and provide an artifact bundle including
generation scripts, corruption prompts, metadata schema, and evaluation code.
\bench is intended as a controlled stress test for RAG
robustness under KB corruption, not as a general model leaderboard.
\end{abstract}

\section{Introduction}
Retrieval-Augmented Generation (RAG)~\citep{lewis2020rag} grounds large
language models (LLMs) in an external knowledge base (KB), giving them
access to up-to-date information, reducing hallucination, and extending
the effective knowledge of smaller
models~\citep{gao2023rag_survey}. The leverage it provides can be
substantial: in our benchmark, a clean KB raises the accuracy of an
8B-parameter model from $42.3\%$ to $98.6\%$ on a multi-domain evaluation,
a gain no amount of parameter scaling could plausibly deliver. The same
model, however, drops to $27\%$ once those KB documents contain factual
corruption, well below its no-retrieval baseline. The same mechanism
that makes RAG effective also makes it fragile.

This fragility is poorly served by current RAG evaluation. Benchmarks
typically assume the retrieval corpus is clean, but real-world KBs
accumulate errors long before any adversary intervenes---a data-quality
problem long studied in the knowledge discovery community: data-entry
mistakes, document versioning conflicts, OCR noise in ingestion
pipelines, outdated content that has not been pruned. At generation
time the model implicitly arbitrates between its parametric prior and
the retrieved context, and the quality of that arbitration decides
whether corrupted documents are discounted or faithfully reproduced.
How much this matters, for which corruption types, and how it should
be evaluated are still open questions. Prior work on RAG robustness
splits into adversarial poisoning, knowledge conflict, and noisy
retrieval, reviewed in Section~\ref{sec:related_work}; none treats
corruption type and severity as joint evaluation variables on a fixed
retrieval pipeline.

We address this gap with \bench, a controlled evaluation
benchmark for stress-testing RAG under systematic KB corruption. The
benchmark specifies four naturalistic corruption types (factual
corruption, numeric typo, relevance poisoning, contradiction
injection) applied at three severity levels (subtle, moderate,
obvious), and a metadata-filtered design in which each of fifteen
experimental conditions is realised as a query-time filter on a single
vector index, so that observed differences reflect KB content and model
arbitration rather than index construction. We instantiate the
benchmark on 57 MMLU subjects~\citep{hendrycks2021mmlu} with a KB of
$182{,}546$ documents (one clean version and twelve corrupted variants
per item, produced by a single LLM-based perturbation pipeline), and
evaluate it on five systems spanning proprietary, reasoning-tuned, and
open-weight tiers: Claude Sonnet~4, Claude Opus~4.6, DeepSeek R1,
Qwen3-32B, and Llama~3.1~8B. The
research questions the design is built to answer, which corruption
types matter (RQ1), whether a model's raw capability predicts its
robustness under corrupted retrieval (RQ2), and whether severity acts
monotonically (RQ3), are formalised as hypotheses
\textbf{H1}--\textbf{H3} in Section~\ref{sec:problem} and tested in
Section~\ref{sec:experiments}.

\paragraph{Contributions.}
We contribute (i)~a controlled KB-corruption evaluation benchmark for
RAG, with clean, no-retrieval, and unfiltered-mixed controls alongside
the fifteen type--severity conditions; (ii)~a single-KB,
metadata-filtered experimental design that isolates corruption type and
severity while holding the retriever and index fixed; and (iii)~a
multi-system evaluation that identifies corruption-specific failure
modes and documents heterogeneity a single-system study would miss. The
artifact---document-construction scripts, corruption-generation
prompts, source questions, and the evaluation and scoring
code---is publicly available,\footnote{\url{https://github.com/Salliberg/CorruptRAG-Code}}
with the evaluation split, metadata schema, document samples, and full
KB pledged for DOI-archived release upon publication (Appendix~F.8).

\section{Related Work}
\label{sec:related_work}

\paragraph{RAG and its evaluation.}
RAG~\citep{lewis2020rag} augments an LLM with a non-parametric document
index, typically retrieved by dense
search~\citep{karpukhin2020dpr}, and successive
surveys~\citep{gao2023rag_survey} identify retrieval quality as the
dominant driver of downstream accuracy.
Retrieval grounds generation partly to curb hallucination, a pervasive
LLM failure mode~\citep{huang2023hallucination}; yet
almost all RAG evaluations assume the retrieval corpus is
clean. A growing line of work tries to close that gap, and our
benchmark sits in a specific corner of it: naturalistic KB corruption,
treated as an evaluation variable rather than an attack objective.

\paragraph{Knowledge-base poisoning and RAG security.}
The most active strand of prior work studies adversarial attacks on
the KB. PoisonedRAG~\citep{zou2024poisonedrag} injects optimised
passages and achieves more than $90\%$ attack success;
Phantom~\citep{shafran2024phantom} uses trigger-based backdoors;
BadRAG~\citep{xue2024badrag} shows that a small number of adversarial
passages suffices; \citet{zhong2023poisoning} inject adversarial passages
that transfer across queries; the single-injection attack of \citet{zhang2025corruptragattack}
improves stealth and feasibility; \citet{zhang2025benchmarking} benchmark 13 attacks
against 7 defences; and RobustRAG~\citep{chaudhari2024robustrag}
provides certifiable robustness via an isolate-then-aggregate defence.
As evaluation protocols, these works adopt the attacker's perspective
and treat corruption as binary (clean or poisoned), optimised for
worst-case attack success rather than the naturalistic, non-adversarial
KB degradation practitioners encounter; our benchmark instead treats
corruption type and severity as controlled \emph{evaluation} variables.
For a broader map of this threat-and-defence landscape see the recent
survey of \citet{mu2026towards}.
Table~\ref{tab:related_work_comparison}
(Appendix~\ref{sec:related_work_comparison}) positions \bench
against the closest prior work on these axes.

\paragraph{Knowledge conflicts and context faithfulness.}
A second strand examines how LLMs arbitrate when retrieved context
contradicts parametric memory~\citep{xu2024knowledge_conflicts}.
\citet{longpre2021entity} find that models tend to over-rely on
parametric memory under entity-substitution counterfactuals, and
\citet{xie2023chameleon} show that models can be either
context-following or prior-stubborn depending on how the conflict is
framed; both studies use narrow entity substitutions on
earlier-generation models. \citet{neeman2023disentqa} report that
training models to predict parametric and contextual answers
separately improves robustness to such conflicts, and
\citet{shi2023irrelevant} demonstrate that LLMs are easily distracted
by irrelevant context in arithmetic reasoning, early evidence that
context utility, not only content truthfulness, affects downstream
accuracy and motivates the semantic-fidelity vs.\ signal-utility
distinction we formalise in Section~\ref{sec:problem}.
\citet{instruction_finetuning2024} further show that instruction
finetuning can \emph{worsen} a model's reliance on context, evidence that
context--memory arbitration is a trained behaviour that varies by system,
consistent with the cross-model heterogeneity we observe.
ClashEval~\citep{wu2024clasheval} extends this line to graded
context vs. parametric conflicts on current-generation models and is
closest in spirit to our severity axis; our benchmark revisits the
same questions with a broader corruption taxonomy.

\paragraph{Robustness of RAG to noisy retrieval.}
A third strand studies what happens when retrieval simply returns
low-quality or off-topic documents. \citet{zeng2025worse} show that
misleading retrievals can push RAG below its no-retrieval baseline;
\citet{rag_noise_analysis2024} find that not all noise degrades
performance, and some noise types can even help; and
\citet{cuconasu2024power} report that irrelevant documents can improve
generation in the right positions. NoMIRACL~\citep{thakur2024nomiracl}
evaluates hallucination under non-relevant retrieval across 18
languages, and \citet{enterprise_robustness2026} measure LLM
perturbation consistency across formats and languages in enterprise
settings. These studies characterise what noise does, but do not vary
KB-corruption type and severity jointly on a common retrieval pipeline.

\paragraph{RAG evaluation benchmarks.}
Existing benchmarks address the evaluation question from different
angles. RGB~\citep{chen2024rgb} evaluates four RAG abilities (noise
robustness, negative rejection, information integration, and
counterfactual robustness); CRAG~\citep{yang2024crag} offers a
large-scale factual-QA benchmark spanning diverse domains and question
types; and RAGAS~\citep{es2024ragas} provides component-level
reference-free metrics for faithfulness, answer relevance, and context
relevance. Closest in structure is QE-RAG~\citep{zhang2025qerag}, which
crosses error types with severity rates, but it perturbs the
\emph{query} rather than the knowledge base, so it measures robustness
to user-side input noise rather than to KB degradation. None of these
treats KB-corruption type, severity, and mixed- versus type-filtered
retrieval as jointly varied evaluation variables in a single controlled
design. Our primary contribution is of this evaluation-design kind,
complementing rather than competing with the attack, defence, and
metric literatures surveyed above.

\section{Problem Formulation}
\label{sec:problem}
\sloppy

We now give a compact formalisation of the evaluation problem. Consider
a language model $\mathcal{M}_\theta$ with parametric knowledge $\theta$
and a knowledge base $\mathcal{K} = \{d_1, \dots, d_N\}$. Given a query
$q$, a retriever $R$ returns the top-$k$ documents
$\mathcal{D}_q = R(q, \mathcal{K})$, and the model produces an answer by
implicitly arbitrating between two information sources,
\begin{equation}
  a \;=\; \mathcal{M}_\theta(q, \mathcal{D}_q)
    \;=\; f\!\left(\underbrace{p_\theta(a \mid q)}_{\text{parametric prior}},\;
                   \underbrace{p(a \mid q, \mathcal{D}_q)}_{\text{contextual evidence}}\right),
  \label{eq:arbitration}
\end{equation}
where $f$ is the implicit arbitration function the model has acquired
during training. When $\mathcal{D}_q$ is accurate, context reinforces
or corrects the parametric prior and accuracy improves; when it is
corrupted, $f$ decides whether the model follows the corruption or
overrides it. We treat $f$ as a behavioural abstraction rather than a
claim about mechanism: what our benchmark measures is how much
observed accuracy changes when $\mathcal{D}_q$ is perturbed in
controlled ways, not the internals of the arbitration.

We model corruption as a function $C_{t,s}: d \mapsto \tilde{d}$ that
perturbs a clean document along one of two information dimensions at
severity $s \in \{\text{subtle}, \text{moderate}, \text{obvious}\}$.
Perturbations along \emph{semantic fidelity} alter the document's
factual claims: \textsc{factual corruption} replaces the correct
answer with a plausible incorrect one while preserving structure and
authoritative tone, and \textsc{contradiction} asserts a wrong answer
to create a direct context--memory conflict. Perturbations along
\emph{signal utility} leave truth intact but reduce the amount of
useful information in the document: \textsc{numeric typo} perturbs
numeric values while preserving the correct answer label, and
\textsc{relevance poisoning} replaces content with topically related
but uninformative filler. This taxonomy has a clear a priori
prediction. Semantic-fidelity perturbations actively mislead, so a
faithful reader must override the context to answer correctly; signal-utility perturbations merely withhold information, and a reader can
fall back on the parametric prior. Semantic-fidelity corruption
should therefore be strictly more harmful than signal-utility
corruption.

Two quantities make this precise and will be used throughout the
paper. The \emph{corruption impact} of type $t$ and severity $s$ on
model $\mathcal{M}$ is the drop in accuracy relative to the clean-KB
ceiling,
\begin{equation}
  \Delta(\mathcal{M}, t, s)
    \;=\; \mathrm{Acc}(\mathcal{M}, \textit{clean})
        \;-\; \mathrm{Acc}(\mathcal{M}, t, s),
  \label{eq:impact}
\end{equation}
so $\Delta \approx 0$ indicates that the model is unaffected and
$\Delta$ large indicates severe degradation. The \emph{resilience gap}
between two models under corruption type $t$, averaged over severity,
is
\begin{equation}
  \Gamma_t(\mathcal{M}_1, \mathcal{M}_2)
    \;=\; \Delta(\mathcal{M}_2, t) \;-\; \Delta(\mathcal{M}_1, t),
  \label{eq:gap}
\end{equation}
where $\Delta(\mathcal{M}, t) = \tfrac{1}{|S|} \sum_{s}
\Delta(\mathcal{M}, t, s)$, so a positive $\Gamma_t$ means $\mathcal{M}_2$ is the more vulnerable
system under corruption of type $t$, independent of any baseline
accuracy difference between the two. We additionally track the
\emph{no-retrieval rate} $\rho(\mathcal{M}, t, s)$, the fraction of
queries for which the retriever returns zero documents, as a
diagnostic for corruption-induced retrievability shifts that would
otherwise confound severity effects.

The research questions of Section~1 translate directly into three
testable hypotheses. \textbf{H1} (RQ1) predicts that semantic-fidelity
corruptions are strictly more harmful than signal-utility ones, i.e.,
for every model $\mathcal{M}$,
\begin{equation*}
\begin{gathered}
\Delta(\mathcal{M}, \textsc{fact}) > \Delta(\mathcal{M}, \textsc{num})
\;\;\text{and}\\
\Delta(\mathcal{M}, \textsc{contra}) > \Delta(\mathcal{M}, \textsc{rel}).
\end{gathered}
\end{equation*}
\textbf{H2} (RQ2) predicts that corrupted-retrieval robustness is
\emph{not} fully explained by parametric knowledge: two systems can
have similar or even inverted no-retrieval baselines yet differ
substantially in $\Delta$ under semantic-fidelity corruption, i.e.,
\begin{equation*}
\begin{gathered}
\mathrm{Acc}(\mathcal{M}_2, \textit{baseline}) \geq
\mathrm{Acc}(\mathcal{M}_1, \textit{baseline})\\
\;\not\Rightarrow\;
\Delta(\mathcal{M}_2, t) \leq \Delta(\mathcal{M}_1, t).
\end{gathered}
\end{equation*}
\textbf{H3} (RQ3) predicts that corrupted RAG can fall below no-RAG
accuracy for sufficiently harmful $(t, s)$ and sufficiently weak
$\mathcal{M}$, and that $\Delta(\mathcal{M}, t, s)$ need not increase
monotonically in $s$ within a given type $t$. Section~\ref{sec:experiments}
tests each of these against the benchmark data.

\section{Evaluation Design}
\label{sec:evaluation_design}

\begin{figure*}[t]
\centering
\resizebox{0.92\textwidth}{!}{%
\begin{tikzpicture}[
  >=Stealth,
  font=\sffamily\small,
  box/.style={draw, rounded corners=3pt, minimum height=0.7cm,
              inner sep=5pt, align=center, line width=0.4pt},
  dimbox/.style={box, minimum width=2.8cm, minimum height=0.6cm},
  typebox/.style={box, text width=2.2cm, minimum height=0.6cm, font=\sffamily\footnotesize,
                  inner sep=3pt},
  sevbox/.style={draw, rounded corners=2pt, minimum width=1.4cm, minimum height=0.4cm,
                 inner sep=3pt, font=\sffamily\scriptsize, line width=0.3pt},
  metricbox/.style={box, text width=3.0cm, fill=gray!5, inner sep=4pt, font=\sffamily\footnotesize},
  arr/.style={->, line width=0.5pt, color=black!60},
  darr/.style={->, line width=0.7pt, color=black!80},
  label/.style={font=\sffamily\scriptsize\itshape, text=black!55},
  heading/.style={font=\sffamily\bfseries\small},
  dimheading/.style={font=\sffamily\bfseries\footnotesize},
]

\node[heading] at (0, 5.7) {Context Arbitration in RAG};

\node[box, fill=blue!8, minimum width=1.2cm] (query) at (-6.0, 4.8) {$q$};
\node[label, anchor=south] at (query.north) {query};

\node[box, fill=green!8, minimum width=2.2cm] (kb) at (-3.2, 4.8) {$\mathcal{K}$};
\node[label, anchor=south] at (kb.north) {knowledge base};

\node[box, fill=orange!10, minimum width=2cm] (docs) at (-0.2, 4.8) {$\mathcal{D}_q$};
\node[label, anchor=south] at (docs.north) {retrieved docs};

\node[box, fill=violet!8, minimum width=3.2cm, minimum height=1.2cm] (model) at (3.2, 4.8) {};
\node[font=\sffamily\bfseries\small] at (3.2, 5.15) {$\mathcal{M}_\theta$};
\node[font=\sffamily\scriptsize, text=black!70] at (3.2, 4.6) {$f\bigl(\underbrace{p_\theta(a|q)}_{\text{prior}},\;\underbrace{p(a|q,\mathcal{D}_q)}_{\text{context}}\bigr)$};

\node[box, fill=gray!8, minimum width=1.2cm] (answer) at (6.0, 4.8) {$a$};
\node[label, anchor=south] at (answer.north) {answer};

\draw[arr] (query) -- (kb) node[midway, above, label] {retrieve};
\draw[arr] (kb) -- (docs) node[midway, above, label] {top-$k$};
\draw[arr] (docs) -- (model);
\draw[darr] (model) -- (answer);

\node[font=\sffamily\scriptsize, text=red!70] (corrupt_label) at (-3.2, 3.7) {$C_{t,s}: d \mapsto \tilde{d}$};
\draw[->, line width=0.5pt, color=red!50, dashed] (corrupt_label) -- (kb);

\node[heading] at (0, 2.9) {Corruption Taxonomy};

\node[dimheading, text=red!70] at (-3.9, 2.2) {Semantic Fidelity};
\node[font=\sffamily\scriptsize\itshape, text=red!50] at (-3.9, 1.85) {actively misleads};

\node[dimheading, text=blue!60] at (3.9, 2.2) {Signal Utility};
\node[font=\sffamily\scriptsize\itshape, text=blue!45] at (3.9, 1.85) {withholds information};

\node[typebox, fill=red!6, draw=red!30] (fact) at (-5.2, 1.1) {Factual\\[-1pt]Corruption};
\node[typebox, fill=red!6, draw=red!30] (contra) at (-2.6, 1.1) {Contradiction\\[-1pt]Injection};

\node[typebox, fill=blue!6, draw=blue!25] (numtypo) at (2.6, 1.1) {Numeric\\[-1pt]Typo};
\node[typebox, fill=blue!6, draw=blue!25] (relpoison) at (5.2, 1.1) {Relevance\\[-1pt]Poisoning};

\node[font=\sffamily\bfseries\scriptsize, text=black!50] at (0, 0.2) {Severity};
\foreach \x/\lab/\col in {-1.5/Subtle/black!20, 0/Moderate/black!40, 1.5/Obvious/black!60} {
  \node[sevbox, fill=\col, fill opacity=0.3, draw=\col, text opacity=1] at (\x, -0.25) {\lab};
}
\draw[->, line width=0.4pt, black!40] (-2.5, -0.25) -- (-2.15, -0.25);
\draw[->, line width=0.4pt, black!40] (2.15, -0.25) -- (2.5, -0.25);

\foreach \n in {fact, contra, numtypo, relpoison} {
  \draw[line width=0.3pt, black!20] (\n.south) -- ++(0, -0.2) -- (0, 0.45);
}

\node[font=\sffamily\scriptsize, text=black!40] at (0, 0.45) {$\times$};

\draw[<->, line width=0.6pt, red!40] (-3.5, 0.55) -- node[midway, fill=white, inner sep=1pt, font=\sffamily\scriptsize\itshape, text=red!60] {$\Delta_{\text{sem}} \gg \Delta_{\text{sig}}$\;(H1)} (3.5, 0.55);

\node[heading] at (0, -1.3) {Evaluation};

\node[metricbox] (impact) at (-5.2, -2.2) {%
  \textbf{Corruption Impact}\\[1pt]
  {\scriptsize $\Delta = \text{Acc}_{\text{clean}} - \text{Acc}_{t,s}$}
};

\node[metricbox] (gap) at (0, -2.2) {%
  \textbf{Resilience Gap}\\[1pt]
  {\scriptsize $\Gamma_t = \Delta_{M_2} - \Delta_{M_1}$}
};

\node[metricbox] (design) at (5.2, -2.2) {%
  \textbf{15 Conditions}\\[1pt]
  {\scriptsize 1 base + 1 clean + 12 corr.\ + 1 unfilt.}
};

\draw[arr, dashed] (0, -0.65) -- (impact.north);
\draw[arr, dashed] (0, -0.65) -- (gap.north);
\draw[arr, dashed] (0, -0.65) -- (design.north);

\node[font=\sffamily\scriptsize, text=black!40] at (0, -3.2) {%
  5 models $\times$ 700 questions $\times$ 15 conditions = 52{,}500 evaluations};

\end{tikzpicture}%
}
\caption{The \bench evaluation space spans three axes:
corruption type (four types in two semantic categories), severity (three
levels), and model (five RAG systems). Each of 700 stratified MMLU
questions is evaluated under 15 conditions (baseline, clean, 12 corrupted,
unfiltered), yielding 52{,}500 evaluations. Conditions are realised via
metadata filters at query time, keeping the retriever and index fixed
across all conditions.}
\label{fig:framework}
\end{figure*}
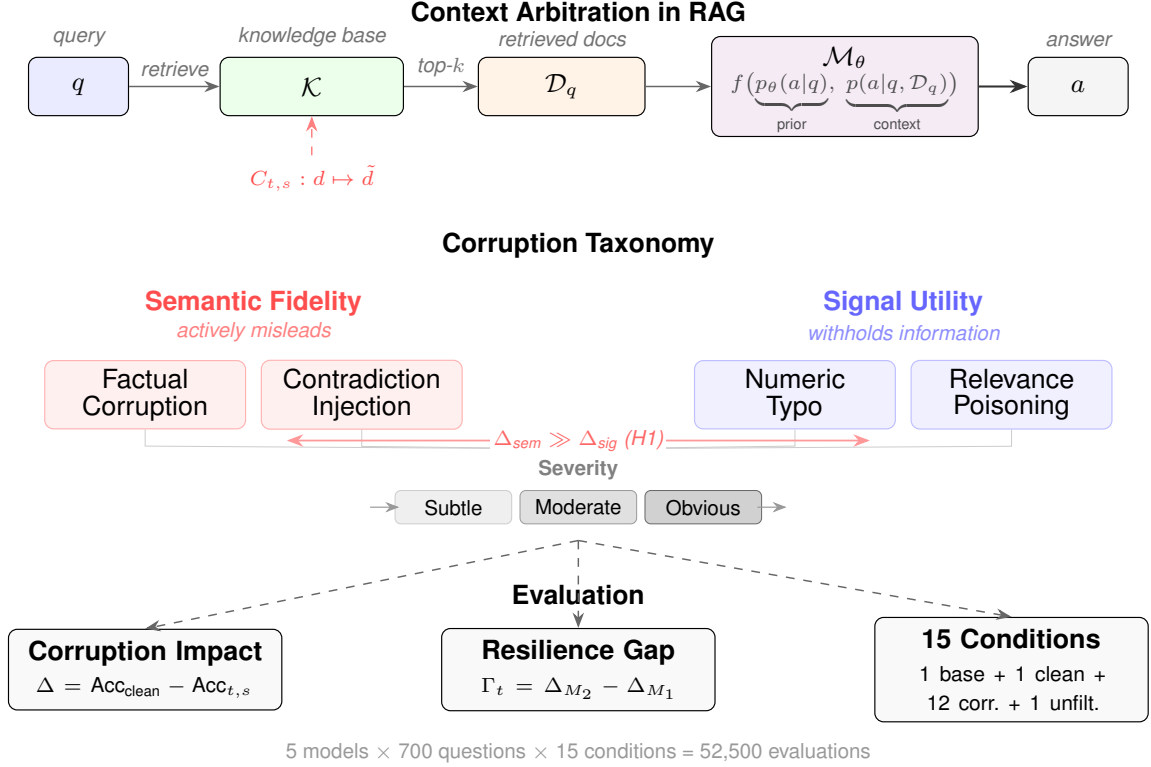

Figure~\ref{fig:framework} summarises the \bench evaluation
framework. A single vector KB stores both clean MMLU-derived reference
documents and their corrupted variants, generated along four corruption
types at three severity levels. Experimental conditions (clean, per-type
$\times$ severity, and unfiltered) are realised at query time via metadata
filters, so every system queries the same index through the same retriever.
The remainder of this section states what the framework is intended to
support, the assumptions it rests on, and how each condition is
implemented.

\subsection{Intended Use and Assumptions}
\label{sec:intended_use}

\paragraph{Intended use.}
\bench is a controlled stress test for RAG robustness under KB
corruption, not a model leaderboard. It supports claims about (i) accuracy
loss under four corruption types, (ii) severity effects, (iii) whether
parametric accuracy predicts corrupted-retrieval robustness, and (iv)
mixed-KB versus type-isolated worst cases. It is not intended for raw
factual-recall comparison, retriever evaluation, or multi-hop or agentic
RAG.

\paragraph{Assumptions.}
The design rests on three assumptions.
\textbf{(A1)} Metadata-filtered retrieval from a single KB is
methodologically equivalent to running each condition against a separate
index, while better controlling for embedding-distribution and
retrieval-score-calibration confounds.
\textbf{(A2)} Exact-match MCQ scoring eliminates LLM-as-judge variance,
attributing differences to KB content; findings may not transfer directly
to open-ended generation.
\textbf{(A3)} The retriever (top-$k{=}5$ dense search) and prompt are fixed
across all conditions and models, so observed differences reflect KB
content and model behaviour, not retrieval or prompt-engineering variation.
We additionally assume that LLM-generated corruptions are representative
of systematic, controlled KB degradation; that document-level chunking is
appropriate for MMLU-length items; and that static, single-turn evaluation
is a lower bound on what systems with active verification could achieve.

\subsection{Single-KB, Metadata-Filter Design}
\label{sec:eval_design}

Maintaining one KB per condition introduces confounds from index
construction, embedding distribution, and document count. Instead, we
build a unified KB containing every document version (clean and all
corrupted variants), each tagged with structured metadata. At query time,
a server-side metadata predicate restricts retrieval to the active
condition's document subset, giving identical retrieval behaviour across
conditions.

\subsection{Experimental Conditions via Metadata Filtering}
\label{sec:conditions}

We define 15 experimental conditions, each enforced by a metadata
predicate at query time. \textbf{Baseline} (no retrieval) measures the
model's parametric knowledge. \textbf{Clean only}
(\texttt{is\_corrupted=false}) establishes the RAG upper bound.
\textbf{Unfiltered} applies no predicate, so corrupted documents outnumber
clean ones 12:1 in the KB and account for ${\approx}79\%$ of retrieved
chunks, simulating an unaudited KB. The remaining \textbf{12 type
$\times$ severity conditions} each restrict retrieval to a single
corruption type and severity level, isolating the effect of each
perturbation class. The full condition table is in
Appendix~\ref{sec:condition_table}.

\section{Benchmark Construction}
\label{sec:benchmark_construction}

\begin{figure*}[t]
\centering
\resizebox{\textwidth}{!}{%
\begin{tikzpicture}[
    >=stealth,
    node distance=0.3cm,
    every node/.style={font=\small},
    stage/.style={rounded corners=5pt, draw=#1!60, fill=#1!8, minimum width=1.8cm, minimum height=1.2cm, align=center, text width=1.6cm, font=\small},
    annot/.style={font=\scriptsize, text=gray!70, align=center},
    arrow/.style={->, thick, color=gray!70},
    bigarrow/.style={->, very thick, color=#1!50},
]

\node[stage=blue] (mmlu) at (0, 0) {MMLU\\[2pt]\footnotesize 57 subjects};
\node[annot, below=0.15cm of mmlu] {14,042 items\\(full test split)};

\draw[arrow] (mmlu.east) -- ++(0.55, 0);
\node[stage=teal] (artgen) at (2.6, 0) {Document\\Generation};
\node[annot, below=0.15cm of artgen] {Claude 3\\Haiku};

\draw[arrow] (artgen.east) -- ++(0.55, 0);
\node[stage=red] (corrupt) at (5.4, 0) {Corruption\\Pipeline};
\node[annot, below=0.15cm of corrupt] {4 types $\times$ 3 sev.\\= 12 variants/doc};

\draw[arrow] (corrupt.east) -- ++(0.55, 0);
\node[stage=violet] (kb) at (8.2, 0) {Unified\\KB};
\node[annot, below=0.15cm of kb] {182,546 docs\\+ metadata};

\draw[arrow] (kb.east) -- ++(0.55, 0);
\node[stage=cyan] (testq) at (11.0, 0) {Test\\Queries};
\node[annot, below=0.15cm of testq] {700 Q from\\raw MMLU};

\draw[arrow] (testq.east) -- ++(0.55, 0);
\node[stage=orange] (retr) at (13.6, 0) {Retrieve\\[2pt]\footnotesize$k{=}5$, filter};
\node[annot, below=0.15cm of retr] {metadata filter\\$\rightarrow$ 15 scenarios};

\draw[arrow] (retr.east) -- ++(0.55, 0);
\node[stage=teal] (models) at (16.2, 0) {5 LLMs};
\node[annot, below=0.15cm of models] {Sonnet, Opus\\DeepSeek, Qwen, Llama};

\draw[arrow] (models.east) -- ++(0.55, 0);
\node[stage=green!60!black] (acc) at (18.8, 0) {Accuracy\\[2pt]\footnotesize vs.\ MMLU GT};
\node[annot, below=0.15cm of acc] {52,500\\evaluations};

\draw[->, thick, dashed, color=violet!50] (kb.south) -- ++(0,-0.9) -| (retr.south);

\node[annot, above=0.4cm of mmlu] (n1) {\textbf{14.0K}};
\node[annot, above=0.4cm of artgen] (n2) {\textbf{14.0K docs}};
\node[annot, above=0.4cm of corrupt] (n3) {\textbf{183K docs}};
\node[annot, above=0.4cm of kb] (n4) {\textbf{183K indexed}};

\draw[bigarrow=blue] (n1.east) -- (n2.west);
\draw[bigarrow=teal] (n2.east) -- (n3.west);
\draw[bigarrow=red] (n3.east) -- (n4.west);

\end{tikzpicture}%
}
\caption{The \bench construction and evaluation pipeline. The
full MMLU test split (14{,}042 items across 57 subjects) is converted into
reference articles by Claude~3~Haiku~\citep{anthropic2024claude3}, each
corrupted into 12 variants (4 types $\times$ 3 severity levels), yielding a
unified 183K-document KB. At evaluation time, 700 questions sampled from
the \emph{original} MMLU are used as test queries; for each query the KB is
searched under 15 metadata-filtered scenarios ($k{=}5$) that control which
corruption condition is retrieved. Responses from five LLMs are validated
against MMLU ground-truth answers, producing 52{,}500 total evaluations.}
\label{fig:pipeline}
\end{figure*}
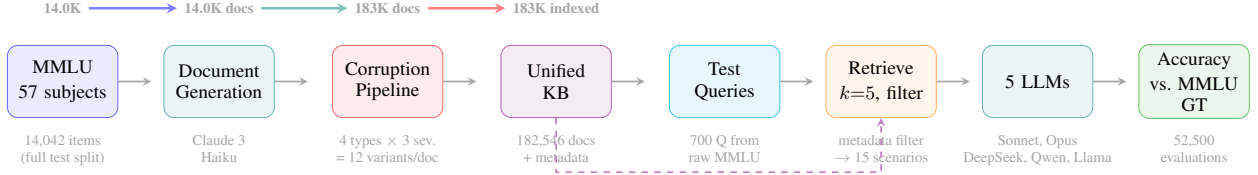

Construction proceeds in three stages: (1) converting MMLU items into
retrievable knowledge documents (the clean corpus); (2) generating
corrupted variants via an LLM-based perturbation pipeline; and (3)
ingesting clean and corrupted documents, with their metadata, into a
single vector KB that supports the conditions in
Section~\ref{sec:conditions}.

\subsection{Dataset: MMLU as a Knowledge-Base Corpus}
\label{sec:dataset}

We ground the study in the \emph{Massive Multitask Language Understanding}
(MMLU) benchmark~\citep{hendrycks2021mmlu}, which spans 57 academic
subjects across STEM, humanities, social sciences, and professional
domains.

\paragraph{Document conversion.}
Each MMLU item (question stem, four choices, correct label) is rewritten
into a short reference-article-style document mirroring an FAQ entry,
keeping the ground truth explicit and machine-verifiable. A representative
example is in Appendix~\ref{sec:doc_example}.

\paragraph{Corpus scale.}
We use the test split of all 57 MMLU subjects, covering STEM (18),
Humanities (13), Social Sciences (12), and Other (14). With one clean and
12 corrupted variants per item (Section~\ref{sec:perturbation}), the full
KB contains \textbf{182{,}546 documents}.

\subsection{Perturbation Method}
\label{sec:perturbation}

\paragraph{Corruption taxonomy.}
We instantiate the four corruption types defined in
Section~\ref{sec:problem}, factual corruption, numeric typo, relevance
poisoning, and contradiction injection, at three severity levels, giving
12 corrupted variants per document. Detailed descriptions and
representative examples are in Appendix~\ref{sec:corruption_details}.

\paragraph{LLM-based generation pipeline.}
All corruptions are generated by Claude~3~Haiku using type-specific prompt
templates. Severity is encoded in the prompt instruction so that levels
produce qualitatively different outputs rather than scaled versions of a
single transformation. The full templates are in
Appendix~\ref{sec:corruption_details} and the artifact bundle.

\paragraph{Metadata tagging.}
Every document is stored with a structured metadata record:
\begin{center}
\small
\setlength{\tabcolsep}{4pt}
\begin{tabular}{@{}lp{0.60\linewidth}@{}}
\toprule
Field & Values \\
\midrule
\texttt{subject}            & MMLU subject name (e.g., \texttt{anatomy}) \\
\texttt{perturbation\_type} & \texttt{clean} $\mid$ \texttt{factual\_corruption} $\mid$ \texttt{numeric\_typo} $\mid$ \texttt{relevance\_poisoning} $\mid$ \texttt{contradiction} \\
\texttt{severity}           & \texttt{none} (clean) $\mid$ \texttt{subtle} $\mid$ \texttt{moderate} $\mid$ \texttt{obvious} \\
\texttt{is\_corrupted}      & \texttt{true} $\mid$ \texttt{false} \\
\texttt{question\_id}       & Unique identifier linking document to its MMLU source item \\
\bottomrule
\end{tabular}
\end{center}

These fields are the mechanism through which experimental conditions are
enforced at query time (Section~\ref{sec:conditions}).

\subsection{Unified Knowledge-Base Ingestion}
\label{sec:kb_construction}

All 182{,}546 documents are ingested into a single vector KB, chunked at
the document level, embedded with a dense model, and indexed with
metadata-filtered approximate nearest-neighbour search. Metadata fields
are stored as indexed attributes so they can drive the predicates that
realise each condition.

\subsection{Representativeness of the Synthetic Corruptions}
\label{sec:representativeness}
Because the corrupted documents are LLM-generated rather than harvested
from a live corpus, we make explicit what real-world degradation each
type is designed to stand in for, and where the analogy stops.
Each corruption type mirrors a distinct, documented failure mode of
production knowledge bases: \emph{factual corruption} models stale or
mis-entered content that remains fluent and authoritative (e.g.\ an
outdated value never pruned after a policy change); \emph{numeric typo}
models data-entry and OCR errors in figures, a dominant error class in
ingested tables and scanned documents; \emph{relevance poisoning} models
versioning drift and topic-adjacent clutter that accumulates when
documents are appended but never curated; and \emph{contradiction
injection} models multi-source KBs that ingest conflicting statements of
the same fact. The severity axis operationalises how conspicuous the
degradation is, from edits a human reader would likely miss to overt
errors. We quantify one facet of representativeness directly: under the
neutral prompt the corruptions do measurably shift model behaviour in
the predicted direction (semantic-fidelity types overturn 25--72\% of
clean-answerable answers; \S\ref{sec:analysis}), confirming they are not
inert artifacts.

Three boundaries of the analogy should be read alongside these mappings,
and are revisited in the limitations of \S\ref{sec:analysis}. First, all
corruptions come from a single generator (Claude~3~Haiku), so absolute
rates may carry a generator signature; because two evaluated systems are
Claude-family, a generator-swap study would isolate this confound, which
we leave to future work. Second, severity is an LLM-assigned operational
label, not a human-validated perceptual scale; we therefore treat
severity as a controlled knob rather than a claim about human-perceived
conspicuousness. Third, documents are derived from benchmark items rather
than a naturally occurring corpus, trading ecological validity for
machine-verifiable ground truth. We accordingly frame \bench as a
\emph{controlled} stress test whose value is comparative (how systems and
corruption types rank relative to one another under identical conditions),
not a forecast of absolute failure rates on any particular deployed KB.

\section{Experiments and Analysis}
\label{sec:experiments}
\newcommand{\ci}[1]{{\scriptsize$\pm$#1}}

\subsection{Experiment Settings}
\label{sec:settings}

\paragraph{Retrieval and generation.}
\label{sec:retrieval_generation}
For each query, the top-$k{=}5$ chunks are retrieved by dense vector
search under the active metadata filter. Retrieved chunks are prepended to
a \emph{neutral} prompt that does not instruct the model to flag
inconsistencies or to distrust context, so any robustness differences
reflect default arbitration behaviours. The response is parsed by
extracting the first letter in $\{A,B,C,D\}$, and accuracy is the exact
match against the MMLU ground-truth label. The answering prompt and the
corruption-generation prompts are described in
Appendix~\ref{sec:corruption_details} and released in the artifact bundle.

\paragraph{Models.}
We evaluate five models spanning proprietary, reasoning-tuned, and
open-weight tiers: Claude Sonnet~4 and Opus~4.6~\citep{anthropic2025claude4,anthropic2026claudeopus46}
(proprietary, integrated retrieve-and-generate, baseline 86.7\% and 78.6\%),
DeepSeek R1~\citep{deepseekr1} (reasoning-tuned hosted, 77.1\%),
Qwen3-32B~\citep{qwen3} (open-weight, retrieve-then-generate, 83.1\%), and
Llama~3.1~8B Instruct~\citep{llama3} (open-weight, 42.3\%). All conditions share
$k{=}5$, the same metadata-filter protocol, and the neutral prompt. Since
serving interfaces differ across systems, we treat exact-match accuracy
as the primary comparable metric.

\paragraph{Evaluation set.}
We use the 700-question stratified sample of
MMLU~\citep{hendrycks2021mmlu} and 15 conditions described in
Sections~\ref{sec:conditions}--\ref{sec:benchmark_construction}, giving
$5 \times 700 \times 15 = 52{,}500$ model--question--condition
evaluations. At $n{=}700$ per cell, Wilson~\citep{wilson1927} 95\%
confidence intervals are
$\pm 3.7$pp at 50\% accuracy and $\pm 1.6$pp at 95\%, sufficient to
detect the 10--71pp effects central to our findings, and feasible within
a one-day API budget.

\paragraph{Hyperparameters.}
Hyperparameters are fixed across runs (Appendix~\ref{sec:hyperparams}):
$k{=}5$ retrieved chunks, document-level chunking, Titan Embeddings V2,
exact-match scoring.

\subsection{Results}
\label{sec:results}

\begin{figure*}[t]
\centering
\includegraphics[width=\textwidth]{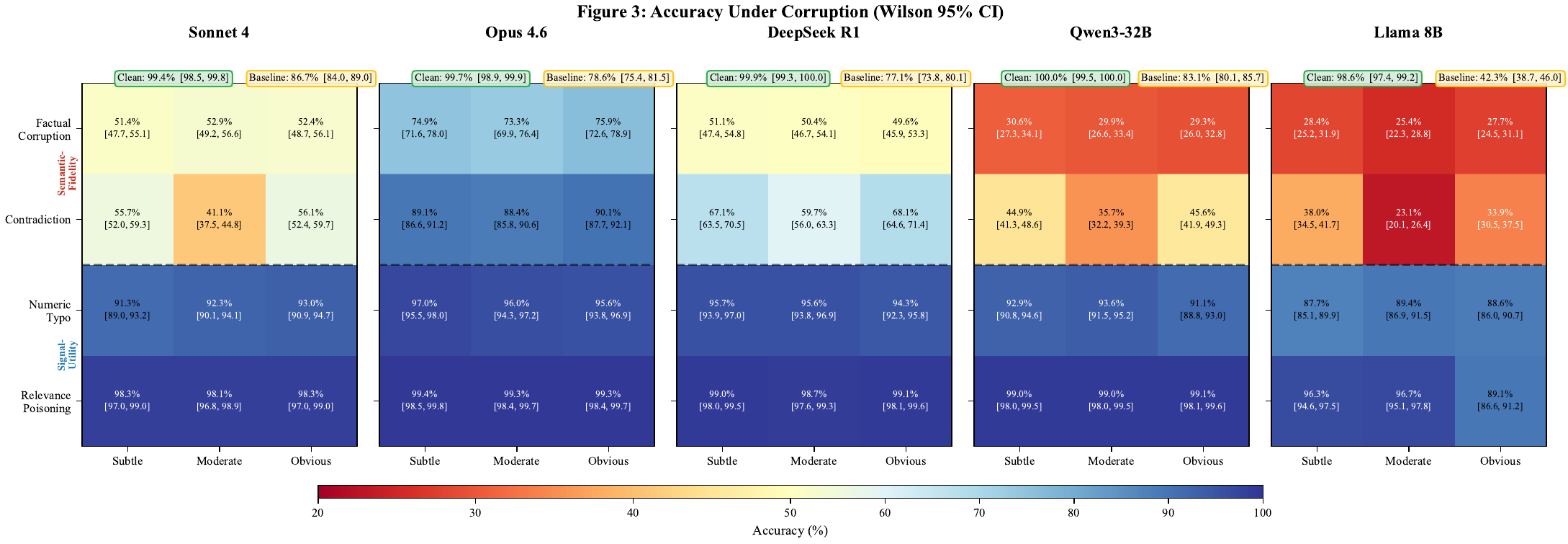}
\caption{Accuracy (\%) with Wilson 95\% CIs across four corruption types,
three severity levels, and five models ($n=700$ per cell). Rows are
grouped into semantic-fidelity (factual corruption, contradiction) and
signal-utility (numeric typo, relevance poisoning) corruptions. The colour
scale is centred at 50\%: red indicates high vulnerability, blue
indicates robustness. Annotation strips show clean-retrieval and
no-retrieval baselines with CIs. Full CI values are in
Table~\ref{tab:full_results}.}
\label{fig:heatmap}
\end{figure*}

Table~\ref{tab:full_results} reports accuracy across all 15 conditions and
Figure~\ref{fig:heatmap} visualises the same data. We report raw accuracy
rather than a single aggregate score, since different corruption types
test different behaviours. A complementary view, the fraction of
clean-answerable questions that each corruption type misleads, is
provided in Figure~\ref{fig:misled_rate} (Appendix).

\begin{table*}[t]
\centering
\caption{Accuracy (\%) with Wilson 95\% confidence intervals across all 15
conditions and 5 models on 700 stratified MMLU questions ($n=700$ per
cell).}
\label{tab:full_results}
\small
\setlength{\tabcolsep}{6pt}
\begin{tabular}{lrrrrr}
\toprule
Condition & Sonnet~4 & Opus~4.6 & DeepSeek R1 & Qwen3-32B & Llama~3.1~8B \\
\midrule
\texttt{baseline\_no\_rag}              & 86.7\ci{2.5} & 78.6\ci{3.0} & 77.1\ci{3.1} & 83.1\ci{2.8} & 42.3\ci{3.7} \\
\texttt{clean\_only}                    & 99.4\ci{0.6} & 99.7\ci{0.4} & 99.9\ci{0.3} & 100.0\ci{0.0} & 98.6\ci{0.9} \\
\texttt{unfiltered}                     & 82.4\ci{2.8} & 95.0\ci{1.6} & 91.3\ci{2.0} & 81.6\ci{2.9} & 74.7\ci{3.2} \\
\midrule
\texttt{factual\_corruption\_subtle}    & 51.4\ci{3.7} & 74.9\ci{3.2} & 51.1\ci{3.7} & 30.6\ci{3.4} & 28.4\ci{3.3} \\
\texttt{factual\_corruption\_moderate}  & 52.9\ci{3.7} & 73.3\ci{3.3} & 50.4\ci{3.7} & 29.9\ci{3.4} & 25.4\ci{3.2} \\
\texttt{factual\_corruption\_obvious}   & 52.4\ci{3.7} & 75.9\ci{3.2} & 49.6\ci{3.7} & 29.3\ci{3.4} & 27.7\ci{3.3} \\
\midrule
\texttt{numeric\_typo\_subtle}          & 91.3\ci{2.1} & 97.0\ci{1.3} & 95.7\ci{1.5} & 92.9\ci{1.9} & 87.7\ci{2.4} \\
\texttt{numeric\_typo\_moderate}        & 92.3\ci{2.0} & 96.0\ci{1.5} & 95.6\ci{1.5} & 93.6\ci{1.8} & 89.4\ci{2.3} \\
\texttt{numeric\_typo\_obvious}         & 93.0\ci{1.9} & 95.6\ci{1.5} & 94.3\ci{1.7} & 91.1\ci{2.1} & 88.6\ci{2.4} \\
\midrule
\texttt{relevance\_poisoning\_subtle}   & 98.3\ci{1.0} & 99.4\ci{0.6} & 99.0\ci{0.7} & 99.0\ci{0.7} & 96.3\ci{1.4} \\
\texttt{relevance\_poisoning\_moderate} & 98.1\ci{1.0} & 99.3\ci{0.6} & 98.7\ci{0.8} & 99.0\ci{0.7} & 96.7\ci{1.3} \\
\texttt{relevance\_poisoning\_obvious}  & 98.3\ci{1.0} & 99.3\ci{0.6} & 99.1\ci{0.7} & 99.1\ci{0.7} & 89.1\ci{2.3} \\
\midrule
\texttt{contradiction\_subtle}          & 55.7\ci{3.7} & 89.1\ci{2.3} & 67.1\ci{3.5} & 44.9\ci{3.7} & 38.0\ci{3.6} \\
\texttt{contradiction\_moderate}        & 41.1\ci{3.6} & 88.4\ci{2.4} & 59.7\ci{3.6} & 35.7\ci{3.5} & 23.1\ci{3.1} \\
\texttt{contradiction\_obvious}         & 56.1\ci{3.7} & 90.1\ci{2.2} & 68.1\ci{3.5} & 45.6\ci{3.7} & 33.9\ci{3.5} \\
\bottomrule
\multicolumn{6}{l}{\footnotesize Values are accuracy (\%) with Wilson 95\% confidence intervals ($n=700$ per cell).}
\end{tabular}
\end{table*}

Averaging over severity (Appendix Table~\ref{tab:type_averages}), factual
corruption and contradiction cause 10--71pp drops, while relevance
poisoning causes drops of $\leq 5$pp.

\subsection{Findings and Analysis}
\label{sec:analysis}

We organise the results around five findings, unified by the theme that
RAG accuracy depends on how the model arbitrates between parametric
knowledge and retrieved evidence, a property that current evaluation
protocols do not isolate.

\paragraph{Finding 1: Clean retrieval saturates accuracy.}
\label{sec:finding1}
\emph{Clean retrieval compresses a 44pp capability gap into a 1.4pp
accuracy band, masking heterogeneity that corruption later reveals.} All
five models reach 98.6--100.0\% with clean context
(Table~\ref{tab:full_results}), despite no-retrieval baselines ranging
from 42.3\% to 86.7\%. The uplift is largest for the weakest model
(Llama~3.1~8B, $+56.3$pp) and sizeable even for mid-baseline systems
(DeepSeek R1, $+22.8$pp from 77.1\%), showing that a clean KB can lift
models at every capability level to a near-ceiling band that is itself
too narrow to distinguish them. The same context-reliance, however,
becomes a failure mode under corruption.

\paragraph{Finding 2: Semantic corruption is far more harmful than utility noise.}
\label{sec:finding2}
\emph{Corruption type matters more than severity: the gap between factual
corruption and relevance poisoning ranges from $\sim$25pp to nearly
70pp across systems, dwarfing the within-type gap between subtle and
obvious variants ($\leq$4pp).}
Factual corruption and contradiction cause 10--71pp drops from clean
retrieval (Appendix Table~\ref{tab:type_averages}), pushing several
systems below their no-retrieval baselines. Relevance poisoning is
comparatively benign ($\leq$5pp drop) for all five systems, suggesting
that off-topic content is easier to discard than fluent misinformation.
Figure~\ref{fig:misled_rate} shows the same pattern in mislead rate:
semantic-fidelity corruptions overturn 25--72\% of clean-answerable
answers, while signal-utility corruptions overturn fewer than 10\%.

\paragraph{Finding 3: No-RAG accuracy does not predict corrupted-retrieval robustness.}
\label{sec:finding3}
\emph{A model with 4.5pp more parametric knowledge can be 45pp less
robust under corruption, indicating that corrupted-retrieval arbitration
is a distinct capability axis.} Qwen3-32B outscores Claude Opus~4.6 at
baseline (83.1\% vs. 78.6\%) yet collapses to 29.9\% under factual
corruption, while Opus stays at 74.7\% (Appendix
Table~\ref{tab:type_averages}). DeepSeek R1 sharpens the same point in
another direction: despite one of the lower baselines (77.1\%), it
reaches near-ceiling clean-KB accuracy (99.9\%) yet suffers a
large semantic-corruption drop (factual corruption collapses to 50.4\%,
$-49.5$pp from clean), on par with Claude Sonnet~4 despite a sizeably
lower parametric baseline. With
five models spanning proprietary, reasoning-tuned, and open-weight
systems this remains illustrative; a broader sweep would test
generality. Quantifying with the resilience gap (Eq.~\ref{eq:gap}):
$\Gamma_{\textsc{fact}}(\text{Opus}, \text{Qwen}) = 45.1$ and
$\Gamma_{\textsc{contra}}(\text{Opus}, \text{Qwen}) = 47.4$. The
robustness divide dwarfs the 4.5pp baseline difference.

\paragraph{Finding 4: Moderate contradictions can be more damaging than obvious ones.}
\label{sec:finding4}
\emph{The most dangerous corruption is not the most extreme one---a
pattern invisible under binary clean-vs.-poisoned protocols.}
Contradiction injection shows a non-monotonic, U-shaped severity pattern
for four of five models (Table~\ref{tab:full_results}): Claude Sonnet~4
drops to 41.1\% at moderate severity, versus 55.7\% (subtle) and 56.1\%
(obvious), a 15.0pp dip whose 95\% CI excludes zero; Qwen3-32B,
Llama~3.1~8B, and DeepSeek R1 (59.7\% vs.\ 67.1/68.1\%) follow the same
shape, though their mod $-$ obv gaps straddle zero in some cases and
warrant further replication. Only Claude Opus~4.6 shows a near-flat
severity profile for contradiction. Severity labels are LLM-assigned and
lack human validation, so this U-shape may partly reflect generator
behaviour; even so, stress tests should include plausible contradictions,
not only extreme ones.

\paragraph{Finding 5: Mixed KBs are not equivalent to worst-case corruption.}
\label{sec:finding5}
\emph{Mixed-contamination robustness and worst-case corruption robustness
are distinct properties; collapsing them reverses system rankings.}
Unfiltered accuracy stays well above type-filtered worst cases for every
model (Appendix Table~\ref{tab:type_averages}): Claude Opus~4.6 drops
only 4.7pp from clean, DeepSeek R1 drops 8.6pp, Claude Sonnet~4 drops
17.0pp, Qwen3-32B drops 18.4pp, and Llama~3.1~8B drops 23.9pp. When
clean evidence is still retrievable, models partially recover, so
unfiltered accuracy should not be reported as worst-case sensitivity.

These findings expose a core tension in RAG: the same context-reliance
that makes RAG powerful makes it fragile. Clean retrieval lifts smaller
models to near-ceiling accuracy, yet semantic corruption erases that
gain and can push performance below random choice. Robustness depends
not on whether the retrieved content is ``noisy'' but on whether the
noise alters the semantic evidence that drives answer selection.

\paragraph{Scope and limitations.}
The benchmark is scoped to MCQ for scoring reliability; open-ended and
multi-hop QA are natural extensions the metadata-filtered design
supports. KB documents are derived from benchmark items, trading
ecological validity for controlled ground truth. All corruptions are
produced by a single generator (Claude~3~Haiku); since two evaluated
models are Claude-family, a generator-swap ablation would quantify the
same-family confound. Severity labels are LLM-assigned without human
validation (see F4 caveat). The evaluated systems use different serving
interfaces, so differences should be read as system-level rather than
purely parameter-level. MMLU items likely appear in the models'
pre-training data, so no-retrieval baselines partly reflect
memorisation; this inflates absolute baselines but leaves the
comparative findings intact, since all conditions share the same
questions and pipeline. Controlled partial corruption (varying the
corrupted fraction in top-$k$) and targeted adversarial injections are
complementary future directions.

\section{Conclusion and Future Work}
\label{sec:conclusion}

We presented a controlled evaluation of how knowledge-base corruption
affects retrieval-augmented generation. Across five models, four
corruption types, and three severity levels, clean retrieval brings every
model close to ceiling accuracy, but corrupted retrieval exposes large
differences in how models arbitrate between retrieved evidence and
parametric knowledge. The dominant failure modes are semantic: factual
corruption and contradiction injection are far more damaging than
numeric typos or relevance poisoning. We further find that no-retrieval
accuracy does not reliably predict corrupted-retrieval robustness, and
that moderate contradictions can be more damaging than obvious ones.
These results argue that RAG evaluation should measure not only whether
retrieval helps under clean conditions, but also whether the model can
reject fluent, relevant, and misleading evidence. The metadata-filtered
framework is reusable: new corruption types, severity levels, or
retrieval configurations can be added without rebuilding the index.

Future work should extend this evaluation in three directions. First,
move beyond MMLU-style multiple choice toward open-ended QA,
citation-heavy generation, multi-hop reasoning, and naturally occurring
enterprise or web documents, in line with broader RAG evaluation
settings~\citep{lewis2020rag,gao2023rag_survey}. Second, study how
retrievers, rerankers, embedding models, and chunking strategies interact
with corruption, building on prior work on noisy and misleading
retrieval~\citep{zeng2025worse}. Third, evaluate defences, retrieval
filtering, contradiction detection, provenance tracking, abstention, and
robust aggregation, under the same controlled protocol, connecting this
benchmark to prior work on RAG poisoning, knowledge conflicts, and robust
RAG defences~\citep{longpre2021entity,zou2024poisonedrag,xue2024badrag,zhang2025benchmarking,chaudhari2024robustrag}.
All artifacts are publicly released or pledged for DOI-archived release
upon publication (Section~1, footnote; Appendix~F.8).

\bibliographystyle{unsrtnat}
\bibliography{reference}

\clearpage
\appendix
\appendix
\section{Positioning Relative to Prior Work}
\label{sec:related_work_comparison}

\begin{table}[h]
\centering
\caption{Positioning of \bench relative to closest prior work on
RAG corruption. Prior work is predominantly attacker-centric or binary
(clean vs.\ poisoned); \bench is the only entry that combines a
controlled corruption taxonomy with graded severity in an evaluation-design
framing.}
\label{tab:related_work_comparison}
\footnotesize
\setlength{\tabcolsep}{3pt}
\begin{tabular}{@{}llll@{}}
\toprule
\textbf{Work} & \textbf{Focus} & \textbf{Corruption} & \textbf{Severity} \\
\midrule
PoisonedRAG~\citep{zou2024poisonedrag}         & Attack    & Adversarial    & Binary \\
Phantom~\citep{shafran2024phantom}             & Attack    & Trigger-based  & Binary \\
BadRAG~\citep{xue2024badrag}                   & Attack    & Adversarial    & Binary \\
RobustRAG~\citep{chaudhari2024robustrag}       & Defense   & Adversarial    & Binary \\
\citet{zhang2025benchmarking}                  & Benchmark & Adversarial    & Binary \\
Worse-than-zero-shot~\citep{zeng2025worse} & Measure & Misleading & Binary \\
RGB~\citep{chen2024rgb}                    & Evaluation & Noisy retrieval & Binary \\
ClashEval~\citep{wu2024clasheval}          & Evaluation & Context conflict & Graded \\
NoMIRACL~\citep{thakur2024nomiracl}        & Evaluation & Non-relevant & Binary \\
\midrule
\textbf{\bench (ours)} & \textbf{Evaluation} & \textbf{Controlled} & \textbf{Graded} \\
\bottomrule
\end{tabular}
\end{table}

\section{Corruption Type Details}
\label{sec:corruption_details}

This appendix expands on the four corruption types and their severity
calibration, supplementing the summary in Section~\ref{sec:perturbation}.
For each type we describe the perturbation, the failure mode it
simulates, and how severity is operationalised in the generation prompt.

\paragraph{Factual corruption.}
The correct answer label is replaced with a different, plausible-but-wrong
option. The question stem and the four choices are left unchanged, so
the document remains internally coherent but asserts a false fact. This
simulates the most dangerous class of KB error: a document that looks
authoritative and well-formed yet contains a wrong answer. Severity
controls the plausibility of the substituted answer---subtle corruptions
choose the most confusable distractor, while obvious corruptions choose a
clearly inferior one.

\paragraph{Numeric typo.}
A realistic transcription error is introduced into a numeric value within
the correct answer, using one of three patterns: a missing digit
($100 \!\to\! 10$), an extra digit ($100 \!\to\! 1000$), or a digit
transposition ($123 \!\to\! 132$). These simulate OCR errors,
copy-paste mistakes, or unit-conversion errors common in automated
ingestion pipelines. Severity controls the magnitude of the introduced
error.

\paragraph{Relevance poisoning.}
Two to three sentences of plausible but off-topic content are appended to
the document. The original correct answer is preserved, so the document
is not factually wrong, but the injected material can distract the
retriever (by shifting the embedding away from the query) or the reader
(by introducing irrelevant context that competes for attention). Severity
controls the topical distance of the injected sentences from the
original subject.

\paragraph{Contradiction injection.}
A contradicting statement is appended that explicitly asserts a different
answer is correct, directly conflicting with the original. Unlike
factual corruption, which silently replaces the answer, contradiction
injection creates an \emph{intra-document conflict}: the document
simultaneously supports two incompatible answers, testing whether the
model can detect and resolve internal inconsistencies in retrieved
context. Severity controls the assertiveness and apparent authority of
the contradicting claim---subtle contradictions hedge with qualifiers,
while obvious ones make bold, unqualified assertions.

\paragraph{Generation and evaluation prompts.}
All corruptions are produced by Claude~3~Haiku from type-specific prompt
templates (one per corruption type, with severity encoded in the prompt
instruction). The evaluator uses a neutral answering prompt that does
not instruct the model to flag inconsistencies or distrust context, so
robustness differences reflect default arbitration behaviour. Full
generation and answering prompt templates, together with the retrieval
configuration, are released in the artifact bundle described in
Sections~\ref{sec:perturbation}--\ref{sec:kb_construction}.

\section{Document Conversion Example}
\label{sec:doc_example}

Each MMLU item is rewritten into a short reference-article-style
document with the following structure (referenced from
Section~\ref{sec:dataset}):

\begin{quote}
\small
\textbf{[Subject] Reference Material}\\[2pt]
\textit{Topic:} \{question stem\}\\[2pt]
According to established knowledge in \{subject\}, the correct answer is
\{label\}) \{correct choice\}.\\[4pt]
Full options for reference:\\
\quad A) \{choice A\} \quad B) \{choice B\} \quad C) \{choice C\}* \quad D) \{choice D\}\\[2pt]
\textit{This information is sourced from authoritative references in \{subject\}.}
\end{quote}

The asterisk marks the correct option in the clean variant. Corrupted
variants share this template, with the perturbation applied to the
\emph{correct answer} field, the option list, or as appended sentences,
depending on corruption type.

\section{Experimental Conditions}
\label{sec:condition_table}

Table~\ref{tab:scenarios} lists the 15 experimental conditions defined
in Section~\ref{sec:conditions}. Each condition is realised by applying
a metadata predicate to the unified KB at query time, so the retriever
and index remain identical across all conditions.

\begin{table}[h]
\centering
\caption{The 15 experimental conditions, each realised as a metadata
filter at query time. ``Type $\times$ severity'' expands to 12 conditions:
$\{\text{factual},\text{numeric},\text{relevance},\text{contradiction}\}\times
\{\text{subtle},\text{moderate},\text{obvious}\}$.}
\label{tab:scenarios}
\small
\setlength{\tabcolsep}{4pt}
\begin{tabular}{@{}l l p{0.30\columnwidth}@{}}
\toprule
\textbf{Condition} & \textbf{Filter} & \textbf{Purpose} \\
\midrule
Baseline (no retr.) & --- & Parametric knowledge reference \\
Clean only & \texttt{is\_corrupted=false} & RAG upper bound \\
Unfiltered & None & Mixed KB (${\approx}79\%$ corrupted retrievals) \\
\midrule
Type $\times$ sev.\ (12) & \texttt{type=$t$, severity=$s$} & Isolate each perturbation class \\
\bottomrule
\end{tabular}
\end{table}

\section{Supplementary Tables and Figures}
\label{sec:supp_tables}
\label{sec:hyperparams}

\begin{figure}[h]
\centering
\includegraphics[width=0.8\linewidth]{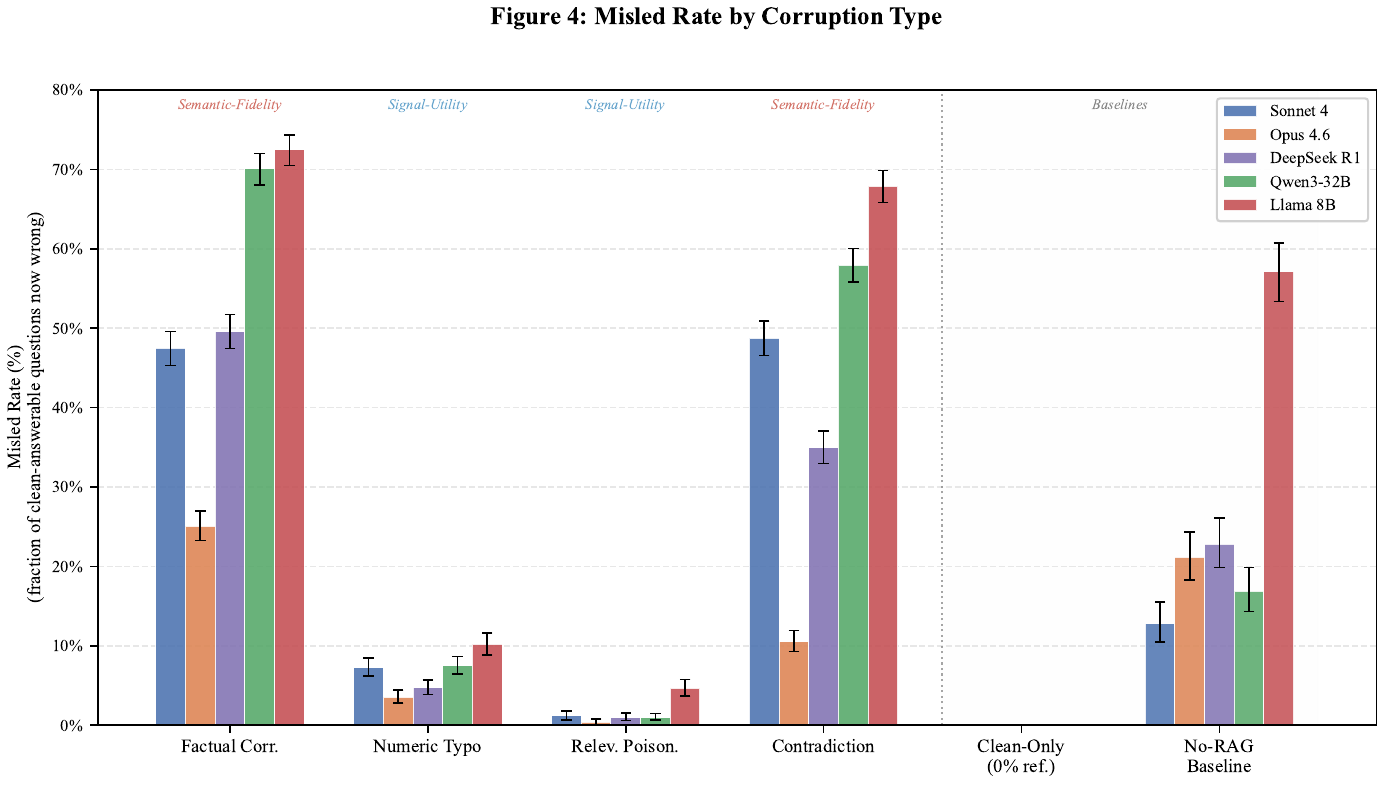}
\caption{Mislead rate by corruption type, defined as the fraction of
questions answerable under clean retrieval that become incorrect under
each corruption type (averaged across severity levels). Semantic-fidelity
corruptions (factual corruption, contradiction) overturn 25--72\% of
clean-answerable answers, whereas signal-utility corruptions (numeric
typo, relevance poisoning) overturn fewer than 10\%. Error bars show
95\% Wilson CIs. This figure is referenced from Finding~2
(Section~\ref{sec:analysis}).}
\label{fig:misled_rate}
\end{figure}

This section collects supplementary material referenced from
Section~\ref{sec:experiments}: the fixed hyperparameters
(Table~\ref{tab:hyperparams}), per-type accuracy averages
(Table~\ref{tab:type_averages}), and a complementary view of corruption
impact via the mislead rate (Figure~\ref{fig:misled_rate}).

\begin{table}[h]
\centering
\caption{Hyperparameters shared across all experimental runs. These
values are held constant so that observed differences across conditions
and models can be attributed to KB content and model behaviour rather
than retrieval or scoring choices.}
\label{tab:hyperparams}
\small
\setlength{\tabcolsep}{4pt}
\begin{tabular}{@{}l p{0.48\columnwidth}@{}}
\toprule
Parameter & Value \\
\midrule
Retrieved chunks per query ($k$)   & 5 \\
Random seed                         & 42 \\
Questions per run                   & 700, stratified across 57 MMLU subjects \\
Embedding model                     & Titan Embeddings V2 \\
Vector index type                   & Approximate nearest-neighbour search \\
Chunking strategy                   & Document-level, one chunk per document \\
Evaluation metric                   & Exact-match accuracy over A/B/C/D \\
\bottomrule
\end{tabular}
\end{table}

\begin{table}[h]
\centering
\caption{Evaluated RAG systems. All systems are accessed through their
providers' standard interfaces and share the retrieval pipeline in
Table~\ref{tab:hyperparams}.}
\label{tab:models}
\small
\setlength{\tabcolsep}{4pt}
\begin{tabular}{@{}lll@{}}
\toprule
System & Provider & Capability tier \\
\midrule
Claude Sonnet~4   & Anthropic         & Proprietary (frontier) \\
Claude Opus~4.6   & Anthropic         & Proprietary (frontier) \\
DeepSeek R1       & DeepSeek (hosted) & Reasoning-tuned \\
Qwen3-32B         & Alibaba           & Open-weight (mid) \\
Llama~3.1~8B Inst. & Meta          & Open-weight (small) \\
\bottomrule
\end{tabular}
\end{table}

\begin{table*}[t]
\centering
\caption{Per-type accuracy (\%) averaged over the three severity levels,
along with the absolute drop (in percentage points) from each model's
\texttt{clean\_only} accuracy. This is the table referenced by Findings
2, 3, and 5 in Section~\ref{sec:analysis}: factual corruption and
contradiction cause 10--71pp drops, while relevance poisoning and numeric
typos remain within $\sim$10pp of clean retrieval.}
\label{tab:type_averages}
\small
\begin{tabular}{lrrrrr}
\toprule
Condition & Sonnet~4 & Opus~4.6 & DeepSeek R1 & Qwen3-32B & Llama~3.1~8B \\
\midrule
\texttt{baseline\_no\_rag}   & 86.7 ($-$12.7) & 78.6 ($-$21.1) & 77.1 ($-$22.8) & 83.1 ($-$16.9) & 42.3 ($-$56.3) \\
\texttt{clean\_only}         & 99.4           & 99.7           & 99.9           & 100.0          & 98.6 \\
\texttt{unfiltered}          & 82.4 ($-$17.0) & 95.0 ($-$4.7)  & 91.3 ($-$8.6)  & 81.6 ($-$18.4) & 74.7 ($-$23.9) \\
\midrule
Factual corruption           & 52.2 ($-$47.2) & 74.7 ($-$25.0) & 50.4 ($-$49.5) & 29.9 ($-$70.1) & 27.2 ($-$71.4) \\
Numeric typo                 & 92.2 ($-$7.2)  & 96.2 ($-$3.5)  & 95.2 ($-$4.7)  & 92.5 ($-$7.5)  & 88.6 ($-$10.0) \\
Relevance poisoning          & 98.2 ($-$1.2)  & 99.3 ($-$0.4)  & 98.9 ($-$1.0)  & 99.0 ($-$1.0)  & 94.0 ($-$4.6) \\
Contradiction                & 51.0 ($-$48.4) & 89.2 ($-$10.5) & 65.0 ($-$34.9) & 42.1 ($-$57.9) & 31.7 ($-$66.9) \\
\bottomrule
\end{tabular}
\end{table*}

\section{\textsc{RAGStress} Benchmark Card}
\label{sec:benchmark_card}

\subsection{Overview}

\setlength{\tabcolsep}{4pt}
\begin{tabular}{@{}l p{0.58\columnwidth}@{}}
\toprule
\textbf{Field} & \textbf{Value} \\
\midrule
Benchmark name        & RAGStress \\
Version               & 1.0 \\
Task type             & Multiple-choice QA under KB corruption \\
Domain                & Multi-domain (57 MMLU subjects) \\
Languages             & English \\
License               & CC-BY-4.0 (benchmark); MIT (code) \\
Total documents       & 182{,}546 \\
Evaluation questions  & 700 (stratified across 57 subjects) \\
Conditions            & 15 (1 baseline + 1 clean + 1 unfiltered + 12 corruption) \\
Corruption types      & 4 (factual corruption, numeric typo, relevance poisoning, contradiction injection) \\
Severity levels       & 3 (subtle, moderate, obvious) \\
\bottomrule
\end{tabular}

\subsection{Intended Use}

\paragraph{Primary use cases:}
\begin{itemize}
  \item Stress-testing RAG systems under controlled KB corruption
  \item Comparing robustness profiles across different model architectures
  \item Evaluating the effect of corruption type and severity on RAG accuracy
  \item Informing KB quality-assurance priorities (which corruption types matter most)
\end{itemize}

\paragraph{Not intended for:}
\begin{itemize}
  \item General model capability ranking or leaderboarding
  \item Evaluating retriever quality in isolation
  \item Open-ended generation quality assessment
  \item Multi-hop, conversational, or agentic RAG evaluation
  \item Measuring real-world attack success rates
\end{itemize}

\subsection{Dataset Composition}

\paragraph{Source data.}
All documents are derived from the MMLU test split~\citep{hendrycks2021mmlu},
which is publicly available under the MIT license. No personal,
sensitive, or copyrighted content is included beyond the original MMLU
items.

\paragraph{Document structure.}
Each source MMLU item produces 13 documents: 1 clean and 12 corrupted
variants (4 types $\times$ 3 severity levels). Documents are short
($\sim$100--200 tokens) reference-article-style passages with structured
metadata (Section~\ref{sec:perturbation}).

\paragraph{Corruption generation.}
Corrupted documents are generated by Claude~3~Haiku using type-specific
prompt templates. The pipeline is deterministic given the same random
seed and model version.

\paragraph{Statistics.}
Per-category counts of source items and resulting documents are shown
below.

\begin{center}
\small
\begin{tabular}{lrrr}
\toprule
Category & Subjects & Source items & Total documents \\
\midrule
STEM            & 18 & 3{,}018 & 39{,}234 \\
Humanities      & 13 & 4{,}705 & 61{,}165 \\
Social Sciences & 12 & 3{,}077 & 40{,}001 \\
Other           & 14 & 3{,}242 & 42{,}146 \\
\midrule
\textbf{Total}  & 57 & 14{,}042 & 182{,}546 \\
\bottomrule
\end{tabular}
\end{center}

\subsection{Collection Process}

\begin{enumerate}
  \item \textbf{Source selection:} MMLU test split, all 57 subjects
  \item \textbf{Document conversion:} automated template-based conversion of MMLU items to reference articles (Appendix~\ref{sec:doc_example})
  \item \textbf{Corruption generation:} LLM-based perturbation (Claude~3~Haiku) with type- and severity-specific prompts
  \item \textbf{Quality filtering:} documents that fail format validation are regenerated (rejection rate $<2\%$)
  \item \textbf{Metadata tagging:} automated assignment of \texttt{subject}, \texttt{perturbation\_type}, \texttt{severity}, \texttt{is\_corrupted}, \texttt{question\_id}
  \item \textbf{KB ingestion:} all documents embedded with Titan Embeddings V2 into a single vector store
\end{enumerate}

\subsection{Evaluation Protocol}

\paragraph{Metrics.} Exact-match accuracy on the four-option multiple-choice answer (A/B/C/D).

\paragraph{Splits.} The 700-question evaluation split is fixed (random seed 42, stratified across the 57 subjects). No training or validation split is provided, since \bench{} is an evaluation-only benchmark.

\paragraph{Baselines.} Each model is evaluated under (1) no retrieval, (2) clean-only retrieval, (3) unfiltered retrieval, and (4) the 12 type $\times$ severity corruption conditions.

\subsection{Maintenance and Versioning}

\begin{itemize}
  \item The benchmark is versioned (current: v1.0). Future versions may add open-ended QA tasks, additional corruption types, or larger question sets.
  \item The source questions, corruption-generation code, document-construction scripts, and evaluation code are publicly available at \url{https://github.com/Salliberg/CorruptRAG-Code}; the evaluation split, metadata schema, corrupted-document samples, and full KB will be archived with a persistent DOI via Zenodo upon publication.
  \item Bug reports and community contributions are tracked via GitHub Issues.
\end{itemize}

\subsection{Ethical Considerations}

\begin{itemize}
  \item The benchmark contains no personal data, hate speech, or harmful content.
  \item Corrupted documents are intentionally wrong; they should not be used as training data without the \texttt{is\_corrupted} metadata flag.
  \item The benchmark is designed for evaluation and stress-testing, not for teaching models to generate misinformation.
  \item We do not release full adversarial prompts that could be directly weaponised for KB-poisoning attacks; we release the corruption \emph{type} descriptions and representative examples.
\end{itemize}

\subsection{Artifacts Released}

\begin{center}
\footnotesize
\setlength{\tabcolsep}{4pt}
\begin{tabular}{@{}p{0.36\columnwidth} l p{0.26\columnwidth}@{}}
\toprule
Artifact & Format & Access \\
\midrule
Source questions (57 MMLU subjects) & JSON & Public release \\
Corruption generation prompts & Python & Public release \\
Document generation scripts   & Python & Public release \\
Evaluation and scoring code   & Python & Public release \\
700-question evaluation split & JSON & Upon publication \\
Metadata schema               & JSON Schema & Upon publication \\
Corrupted doc.\ sample (1{,}000) & JSON & Upon publication \\
Full KB (182{,}546 docs) & JSON/Parquet & Zenodo DOI upon publication \\
\bottomrule
\end{tabular}
\end{center}

\end{document}